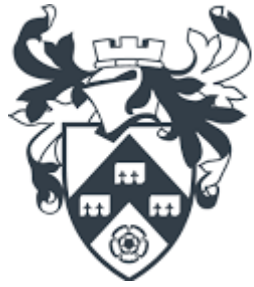

Department of Computer Science

# EVALUATING THE PERFORMANCE OF STATE-OF-THE-ART ESG DOMAIN-SPECIFIC PRE-TRAINED LARGE LANGUAGE MODELS IN TEXT CLASSIFICATION AGAINST EXISTING MODELS AND TRADITIONAL MACHINE LEARNING TECHNIQUES

Module Code: COM00151M
Module Name: Independent Research Project

by Tin Yuet CHUNG
June 2024
supervised by Dr. Majid Latifi

word count: 9395
A dissertation submitted in partial fulfilment of the requirement for
the degree of
Master of Science in Computer Science with Artificial Intelligence

## Acknowledgement

First of all, I would like to thank Dr. Majid Latifi for his supervision and help throughout the duration of the project. In the last few months, Majid has provided me the direction of research and has given insightful comments. I also want to thank Tobias Schimanski and his research group who contributed to the ESG annotated text data.

Secondly, I would like to thank my friends in Hong Kong and Canada.

Last but not least, I would like to thank my family for supporting me in Canada.

# Table of Contents

# List of Figures

# Lists of Tables

# Table of Abbreviations

| Abbreviation | Description |
|---|---|
| AI | Artificial Intelligence |
| BERT | Bidirectional Encoder Representations from Transformers |
| ESG | Environmental, Social, and Governance |
| XAI | Explainable AI |
| XGBoost | Extreme Gradient Boosting |
| GPT | Generative Pre-trained Transformer |
| LLaMA 2 | Large Language Model Meta AI 2 |
| NER | Named Entity Recognition |
| QLoRA | Quantized Low Rank Adapters |
| RNNs | Recurrent Neural Networks |
| SRI | Sustainable Responsible and Impact Investing |
| SVM | Support Vector Machine |
| ULMFiT | Universal Language Model Fine-tuning |
| NF4 | 4-bit Normal Float |
| LLMs | Large language models |
| ML | Machine Learning |
| PEFT | Parameter-Efficient Fine-Tuning |
| RL | Reinforcement Learning |
| SFT | Supervised Fine-Tuning |

# Executive Summary

This research investigates the classification of Environmental, Social, and Governance (ESG) information within textual disclosures. The aim is to develop and evaluate binary classification models capable of accurately identifying and categorizing E, S and G-related content respectively.

The motivation for this research stems from the growing importance of ESG considerations in investment decisions and corporate accountability. Accurate and efficient classification of ESG information is crucial for stakeholders to understand the impact of companies on sustainability and to make informed decisions.

The research uses a quantitative approach involving data collection, data preprocessing, and the development of ESG-focused Large Language Models (LLMs) and traditional machine learning (Support Vector Machines, XGBoost) classifiers. Performance evaluation guides iterative refinement until satisfactory metrics are achieved.

 The research compares traditional machine learning techniques (Support Vector Machines, XGBoost), state-of-the-art language model (FinBERT-ESG) and fine-tuned LLMs like Llama 2, by employing standard Natural Language Processing performance metrics such as accuracy, precision, recall, F1-score. A novel fine-tuning method, Qlora, is applied to LLMs, resulting in significant performance improvements across all ESG domains. The research also develops domain-specific fine-tuned models, such as EnvLlama 2-Qlora, SocLlama 2-Qlora, and GovLlama 2-Qlora, which demonstrate impressive results in ESG text classification.

Key findings show that fine-tuned LLMs, particularly those utilizing Qlora, achieved average F1-score improvements of 7.37% over classical machine learning models; and 12.30% over FinBERT-ESG. However, the high computational demands of LLMs pose a potential barrier for organizations with limited resources. The research highlights the need for exploring more cost-effective alternatives for specific use cases where the exceptional performance of Qlora-fine-tuned models might not be essential.

The deployment of fine-tuned LLMs like those utilizing Qlora presents several issues. Legally, organizations must navigate intellectual property and licensing agreements; socially, the high computational demands can widen the digital divide by favouring well-resourced entities; ethically, the significant energy consumption of these models raises environmental sustainability concerns; and professionally, there's an increased demand for data scientists to acquire skills in optimizing resource-intensive models to stay competitive.

While this research contributes valuable tools and insights to the field of ESG impact classification, it also presents several avenues for future work. Expanding the training data to include diverse ESG-related text sources and exploring data augmentation techniques enhance model robustness and generalizability. Further investigation into alternative fine-tuning methods like RAG and prompt engineering could reveal additional performance improvements. Lastly, utilizing newer and more advanced LLMs like Llama 3 holds potential for even greater accuracy and nuance in ESG classification. This research highlights the significant advancements in natural language processing techniques for ESG information classification. By continuously refining methods and exploring new technologies, future research can contribute to a more sustainable and responsible future.

# 1- Introduction

## 1.1 Motivation and Background

The increasing prevalence of socially responsible investing has elevated Environmental, Social, and Governance (ESG) factors to a position of prominence in the decision-making processes of investors and stakeholders. This emphasis on sustainability necessitates robust tools for evaluating and comprehending corporate ESG performance [1], [2]. However, the sheer volume, diversity, and unstructured nature of ESG data – ranging from extensive corporate reports to social media sentiment – pose significant challenges for traditional machine learning (ML) models, exposing their limitations in processing such complex and dynamic data [3].

Sustainable Responsible and Impact Investing (SRI) calls for transparency and accountability [4], [5]. Investors require accurate predictions and the ability to understand the reasoning behind them – a crucial limitation of traditional "black box" ML techniques [3].

The field of Natural Language Processing (NLP) has experienced a revolution as result of the advance of transformer-based language models [6]–[8]. These powerful models have paved the way for domain adaptation techniques, allowing them to excel in specialized areas such as ESG analysis [9]. Deep learning algorithms such as GPT (Generative Pre-trained Transformer) and BERT (Bidirectional Encoder Representations from Transformers) are particularly adept at understanding the nuances of human language and its context [7], [8], [10], making them invaluable tools for various NLP tasks.

However, such large language models (LLMs) pose their own challenges. Extensive parameter sets demand substantial data, computational resources, energy, and training time. This raises questions about whether their complexity is justified given their potential implementation cost [11]–[13]. Additionally, their opaque nature hinders interpretability, a potential obstacle for applying LLMs in ways that require clear insights into model decisions [14].

## 1.2 Research Aim

This project explores the cutting-edge performance of Large Language Model Meta AI 2 (LLaMA 2) [15], fine-tuned for ESG applications. Its refined architecture and vast training dataset [16], [17] could be particularly valuable for understanding the complexities of ESG reporting. The project aims to have models that accurately categorise ESG texts and outperform classical ML models like Support Vector Machine (SVM) [18] and Extreme Gradient Boosting (XGBoost) [19], as well as state-of-the-art FinBERT-ESG [20]. These would be valuable tools for businesses, investors, and researchers who need to sift through large volumes of text to make informed decisions based on ESG criteria.

## 1.3 Research Focus

Adapting LLaMA 2 [15] from its general knowledge base to the nuances of ESG terminology and concepts is a core challenge. Finding the most effective fine-tuning strategies to instil this specialized understanding is crucial for success. Moreover, determining which advanced fine-tuning techniques, like Quantized Low Rank Adapters (QLoRA) [21], work best with LLaMA 2 for ESG classification, and finding the optimal settings for these techniques, requires significant experimentation. This process must balance potential performance gains with the need to manage computational costs and maintain a degree of model interpretability.

## 1.4 Research Questions/Hypotheses

Two research questions are proposed: (1) How can we develop a fine-tuned language model specifically for the ESG domain? (2) How does the performance of the newly developed ESG LLM compare to existing models on ESG text classification tasks?

This project proposes the development and evaluation of a state-of-the-art ESG-specific LLM. LLaMA 2 could be fine-tuned for ESG text classification, by using QLoRA. By fine-tuning LLaMA 2 for ESG text classification, this LLM could significantly outperform existing models, such as FinBERT-ESG and traditional ML baselines in ESG text classification tasks.

## 1.5 Research Objectives

The research objectives are as follows:

1. Conduct a comprehensive review of existing research on ESG and LLMs. This establishes a strong foundation for the project and identify potential gaps or areas for improvement.
2. Develop a robust research methodology for creating and evaluating an ESG-specific LLM. Focus on defining the approach and procedures to be used in this process.
3. Design and execute rigorous evaluation methods to assess the proposed LLM. This should include clearly defined performance metrics (accuracy, precision, recall, F1-score, etc.).
4. Benchmark the developed LLM against existing models (FinBERT-ESG) and traditional ML approaches (SVM and XGBoost). This determines its relative strengths and any potential enhancements.

## 1.6 Structure of the dissertation

This dissertation is organized as follows:
- Chapter 2 covers the literature review. Fundamentals of this research, such as ESG, LLM, BERT, LLaMA 2, are discussed. Classical ML like XGBoost and SVM, as well as techniques of fine-tuning LLM, are also be discussed. The research gap was highlighted.

- Chapter 3 presents the methodology, design, implementation, and the discussion on computation approach.

- Chapter 4 illustrates results, analysis, evaluation and testing of the fine-tuned state-of-the-art ESG-specific LLM.

- The conclusion was drawn, in Chapter 5, and future work was discussed.

# 2- Literature Review

## 2.1 Environment, Social, and Governance (ESG)

ESG provides a framework for evaluating a company's commitment to sustainability and ethical practices [22]. It encompasses environmental factors like climate change mitigation, waste management, resource conservation, and biodiversity; social factors like human rights, consumer protection, workforce health and safety, and workforce development; and governance factors like board independence, ethical practices, and anti-bribery measures. These multifaceted criteria impact a company's performance and long-term viability [23].

The rise of ESG reflects a growing understanding of the interconnected nature of business and sustainability [24]. Companies are recognizing that their long-term success hinges on proactively addressing environmental, social, and governance concerns. This shift is driven by stakeholders who understand the risks associated with unsustainable practices and the opportunities presented by responsible business conduct [25].

The urgency to address climate change and create a more sustainable business landscape is the primary driver of ESG integration. The consequences of unchecked environmental damage are potentially catastrophic. Transitioning to a sustainable economy prioritizes ESG performance alongside financial returns [26]. This transition may create disruptions as industries adapt to new regulations, consumers shift their purchasing behaviours, and outdated business models become obsolete. In response, regulatory bodies continue to strengthen and formalize ESG reporting and integration requirements [27].

For financial institutions, navigating this evolving landscape requires reliable ESG data and sophisticated analysis [28]. The ability to assess companies and financial products through an ESG lens offers a significant competitive advantage. This allows institutions to identify potential risks, align investments with sustainable goals, attract ethically-minded investors, and demonstrate a commitment to responsible practices [29]. Incorporating ESG principles is not only a matter of social responsibility but a cornerstone of sound financial strategy in a world where sustainability and financial performance are becoming increasingly intertwined [30].

## 2.2 Large Language Model (LLM)

Humans are born communicators, using language to express ourselves since infancy [31], [32]. However, machines historically lacked this ability. The quest for machines that can read, write, and converse like humans has been a longstanding challenge in AI [33].

The advance of deep learning techniques coupled with the availability of powerful computing resources and vast quantities of text data has propelled the development of LLMs [34]. These sophisticated AI models, encompassing billions of parameters, are trained on colossal amounts of unlabelled text data sourced from the internet. Through this self-directed learning process, LLMs are able to comprehend intricate language patterns, subtle nuances, and intricate connections within textual data [35].

LLMs have proven adept at various language tasks, including generating different creative text formats, translation, summarization, question answering, and understanding emotions in text (sentiment analysis) [36]. They achieve this by making use of deep learning and vast datasets. Fine-tuning these models for specific tasks has shown impressive results, even achieving cutting-edge performance on certain benchmarks [37].

LLMs stem from earlier language models and neural networks. Initial attempts used simpler models that weren't well-suited for capturing the complexities of language, especially distant connections and broader picture. With the rise of neural networks and more data, researchers explored more sophisticated approaches [38]. The development of Recurrent Neural Networks (RNNs) [39] was a significant step, allowing for modelling sequential data like language. However, RNNs had limitations in handling long-range dependencies.

A breakthrough came with the transformer architecture. This powerful model uses a mechanism called self-attention [40], enabling efficient processing of long-range dependencies and allowing for parallel processing. Transformer-based architectures like Google's BERT [8] and OpenAI's GPT [10] series have become the foundation for LLMs that excel at various language tasks.

## 2.3 BERT (Bidirectional Encoder Representations from Transformers)

Since its debut in 2018 [8], BERT has emerged as a revolutionary model in the field of NLP. Its architecture is founded upon the Transformer model, which utilizes attention mechanisms to discern and represent the relationships between words within a sentence [41]. BERT's key distinction lies in its bidirectional training approach. This means the model processes text by considering the context from both preceding and succeeding words, a significant advancement from earlier models that were limited to unidirectional analysis. This bidirectional capability enables BERT to develop a more comprehensive understanding of language and context [8].

Despite its groundbreaking impact, BERT has certain disadvantages compared to newer large language models. One significant limitation of BERT is its computational inefficiency and slower inference speed, making it less suitable for real-time applications where rapid responses are crucial [42]. Additionally, BERT struggles with processing long documents due to its input token length restriction, which can hinder its performance in tasks requiring extensive textual data analysis [43].

In response to BERT's limitations, newer large language models like RoBERTa and ALBERT have been developed to address these shortcomings. For instance, ALBERT aims to reduce memory consumption and training time compared to BERT, enhancing efficiency without compromising performance [44]. These models have shown improvements in various NLP tasks, outperforming BERT in certain aspects [45].

Moreover, modifications to the BERT architecture have been proposed to enhance its capabilities. Tasks like semantic similarity search and clustering require sentence embeddings that capture meaning effectively. Sentence-BERT (SBERT), developed by Reimers and Gurevych [43], offers a solution for deriving such embeddings, addressing BERT's limitations in these areas. Additionally, research has explored combining BERT with other architectures, such as CRF, to improve performance in specific tasks like aspect extraction [46].

While BERT has been a transformative model in NLP, newer large language models have emerged to overcome its limitations and enhance efficiency and performance in various tasks [47]. By addressing issues such as computational inefficiency and document length restrictions, these models have pushed the boundaries of NLP, underscoring the continuous evolution and improvement in the field [48].

## 2.4 LLaMA2 (Large Language Model Meta AI 2)

Similar to many other high-performing LLMs, LLaMA 2 is built upon the Transformer architecture, well-known for its self-attention mechanism that enables it to grasp the relationships between words in a sentence [17]. Building upon the foundation laid by its predecessor, the original LLaMA model [49], LLaMA 2 represents a substantial advancement in both language comprehension and generation capabilities. This progress is attributed to its intricate design, the utilization of an extensive training dataset, and the implementation of sophisticated learning techniques during its development [50].

Both it and its predecessor rely on the Transformer model [50], but LLaMA-2's increased size (70B parameters) and greater depth allow it to grasp more complex language patterns [51]. The diversity of LLaMA 2's training data is key: it includes internet sources, academic texts, and more [15], giving it the broad language knowledge needed to excel across multiple languages and subject areas [52].

In comparison to BERT and other LLMs like GPT-3 [53], LLaMA2 offers several advantages. One key benefit is its task-specific approach, enabling it to excel in specialized tasks where general models may fall short [54]. LLaMA2's architecture allows it to capture complex linguistic nuances and context-specific information, leading to improved performance in tasks requiring domain-specific knowledge. Demonstrating its adaptability and potential within specialized domains [55], LLaMA2, in conjunction with GPT-3, has undergone fine-tuning to effectively address inquiries pertaining to technical specifications and documentation. Additionally, LLaMA2 has demonstrated robustness and efficiency in generating accurate results with lower computational costs, making it a practical choice for various NLP applications [15].

In the realm of ESG text categorization, LLaMA2 holds significant promise. Its proficiency in understanding and processing intricate textual data, combined with its task-specific models, can be harnessed to categorize ESG-related texts effectively [56]. By training LLaMA2 on ESG-specific corpora and fine-tuning it for ESG text analysis, organizations can benefit from more precise and efficient categorization of ESG-related content [57].

LLaMA2 emerges as a promising language model with specialized capabilities that offer advantages over traditional models like BERT [58]. Its potential in ESG text categorization underscores its versatility and applicability across diverse domains, highlighting its value in advancing NLP tasks [59].

## 2.5 Classical Machine Learning (ML), Support Vector Machine (SVM) and XGBoost

Classical ML encompasses established data analysis methods distinct from deep learning techniques. These methods involve algorithms that discern patterns and generate predictions from data without the need for explicit programming for each specific task [60]. Support Vector Machines (SVMs), a prime example of classical ML algorithms, excel in classification and regression tasks by identifying the optimal hyperplane that effectively separates different classes within the feature space [61]. In contrast, XGBoost, a robust implementation of the gradient boosting framework, is widely employed for both regression and classification problems. It operates by creating a sequence of decision trees, each iteratively refining the errors of its predecessors, resulting in highly accurate predictions [62].

Within the domain of ESG text categorization, ML techniques play a pivotal role in analysing and classifying textual data related to sustainability and corporate responsibility initiatives [63]. A research study conducted by Jing [64] investigated the correlation between a company's financial performance and ESG disclosure scores. Employing regression models, the study established relationships between these factors, showcasing the application of traditional ML methods in comprehending the influence of ESG practices on financial results.

Moreover, when it comes to uncovering corporate sustainability challenges raised in shareholder resolutions, machine learning-powered text analysis techniques have emerged as powerful tools. By delving into the unstructured text data of these resolutions, these methods extract valuable insights [65], essentially converting raw textual information into actionable knowledge. This capability significantly aids in recognizing and comprehending sustainability-related concerns that exist within organizations [66].

Furthermore, the application of ML in sentiment analysis, as shown in the study by Wadhwani et al. [67], illustrates how supervised ML models can be leveraged to analyse sentiment in Twitter data related to events such as the Russia-Ukraine war. This application underscores the versatility of classical ML algorithms in processing and categorizing text data to extract meaningful information regarding public sentiment and opinions [68].

Classical ML algorithms such as SVM and XGBoost are essential in ESG text categorization tasks, empowering the discovery of key knowledge hidden within textual information related to environmental, social, and governance aspects [69]. These algorithms are pivotal in analysing ESG disclosures, identifying sustainability issues, and conducting sentiment analysis, thereby contributing to a deeper understanding of the implications of ESG practices on various domains [70].

## 2.6 LLM Fine-Tuning

### 2.6.1 What is LLM Fine-Tuning?

Fine-tuning is a powerful technique employed to adapt pre-trained models, both in the realm of LLMs and traditional ML, for specific tasks or domains. Rather than building a model from the ground up, which demands significant computational resources and time [71], fine-tuning leverages the knowledge base of a pre-trained model and adjusts its parameters to suit the new task. This procedure involves updating a portion of the model's weights according to the new data or examples relevant to the target task [72].

This approach offers several advantages. Firstly, it significantly reduces the time and computational resources needed, in contrast to training a model from zero. By utilizing a pre-trained model as an initial basis, the starting training phases are bypassed, leading to faster achievement of optimal results [73]. Secondly, fine-tuning is a key component of transfer learning, where knowledge gained from one task is moved to another. This is particularly useful when handling limited labelled data for the new task, as the model can draw on its prior understanding to learn effectively [74].

Moreover, fine-tuning allows for customization, enabling the tailoring of a pre-trained model for highly specific tasks such as sentiment analysis or generating text within specialized domains like healthcare or customer service [75]. It also provides adaptability, allowing continuous adjustments to the model as data evolves or needs change, eliminating the need for retraining from scratch [76]. Additionally, fine-tuning can help mitigate potential biases present in pre-trained models by using balanced and representative data during the process [77]. Finally, for sensitive data subject to privacy regulations, fine-tuning allows customization within a secure environment, ensuring data protection [78].

## 2.6.2 Advantages and real-world applications of fine-tuned models

For advanced control over NLP projects, fine-tuning a language model can offer substantial gains in several areas. Fine-tuning helps a model more accurately detect and categorize subjective information within text, improving its ability to interpret human emotions and opinions in sentiment analysis tasks [79]. Within Named Entity Recognition (NER), fine-tuning makes the model more adept at identifying and labelling key entities (e.g., people, places, organizations) [80]. For text generation applications, fine-tuning tailors the model's output for specific styles, tones, or formats, enhancing its use in creative writing assistance and advanced chatbots [81]. Additionally, fine-tuning for particular language pairs significantly improves the quality and fluency of machine translation systems [82]. A fine-tuned language model can also produce clearer, more concise summaries of longer texts, aiding content management tasks [83], and it powers more accurate context-based question answering systems [84]. Finally, a fine-tuned model strengthens conversational agents or chatbots through a focused understanding of context, enabling a smoother and more relevant user experience [85].

## 2.6.3 LLM Fine-Tuning Techniques

LLM fine-tuning techniques encompass a spectrum of approaches, from classic methods to modern innovations, each designed to tailor these powerful models for specific applications [86].

Classic approaches centre around several methods. Feature-based techniques leverage the pre-trained LLM as a feature extractor, converting text input into a fixed-size vector. A separate classifier then learns to predict categories based on this representation. While computationally efficient, this approach can sacrifice some performance accuracy [87]. Fine-tuning I (by updating the output layer) enhances this by adding extra layers on top of the pre-trained LLM, with only these newly added layers updated during training. This often yields slightly better results than the feature-based approach. Alternatively, Fine-tuning II (by updating all layers) involves unfreezing the entire model, including the pre-trained LLM, allowing all parameters to be adjusted; however, this risks "catastrophic forgetting" where new learning overwrites prior knowledge. Though resource-intensive, it offers the best performance when this is the top priority [88].

Beyond these, techniques like ULMFiT (Universal Language Model Fine-tuning) provide a transfer learning approach for NLP, using a specialized architecture for fine-tuning [89]. Additionally, both gradient-based [90] and Random Forest-based parameter importance ranking [91] methods assist in identifying the most crucial features or parameters within a model – essential for optimization.

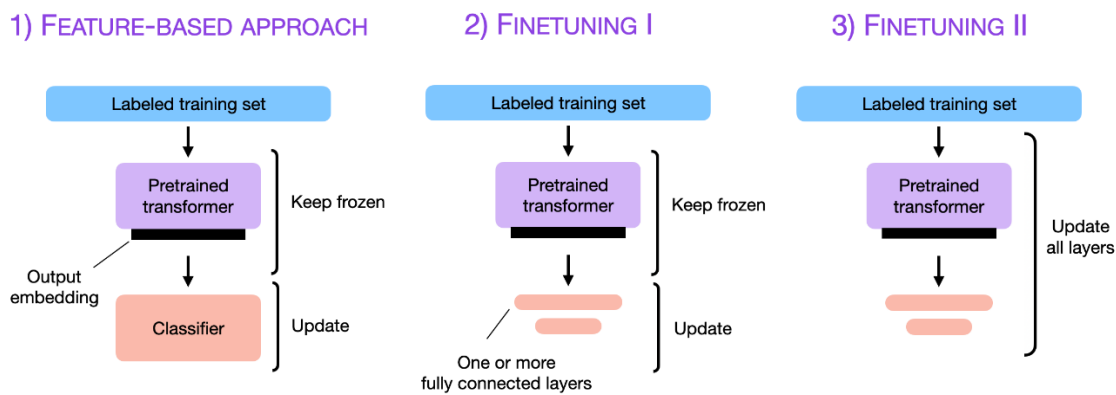

*Figure 1: classic approaches of LLM fine-tuning "Adapted from: [88]*

### 2.6.4 Qlora

Traditional fine-tuning of LLMs involves adjusting all the weights within the model's vast network [92]. LoRA [93], or Low-Rank Adaptation, presents an alternative approach that significantly reduces the computational burden and memory requirements of this process. Instead of modifying the entire weight matrix of the pre-trained LLM, LoRA introduces the concept of "adapters" [94]. These adapters are composed of two smaller matrices that effectively resemble the larger original matrix. The fine-tuning process focuses solely on adjusting the weights within these smaller matrices, leaving the pre-trained model's weights untouched. Once the adapter is fine-tuned, it is seamlessly integrated with the pre-trained model and employed during inference [93]. This efficient approach allows for adaptation of LLMs to specific tasks while minimizing computational overhead and memory consumption [95].

Taking inspiration from LoRA's efficient fine-tuning approach, QLoRA [21] introduces further optimizations to minimize memory consumption without sacrificing performance. The core principle behind QLoRA's memory savings lies in 4-bit quantization of the pre-trained LLM's weights. This contrasts with the 8-bit quantization typically used in LoRA [96], and it allows for a significant reduction in the model's memory footprint, both when it is being trained and when it is actively making predictions. Despite this lower precision, QLoRA maintains a level of effectiveness similar to that of LoRA [21].

QLoRA's impressive memory efficiency stems from a combination of innovative techniques that work synergistically to minimize memory consumption during LLM fine-tuning. One crucial aspect is the utilization of 4-bit NormalFloat (NF4), a novel data type specifically tailored for normally distributed weights [97]. NF4 ensures optimal information representation within a compact 4-bit format, significantly reducing the memory required to store the model's weights. Furthermore, QLoRA extends the concept of quantization beyond just the model weights. It also applies quantization to the quantization constants themselves through a process known as double quantization [98]. This further shrinks the memory footprint by representing these constants with fewer bits. To address potential memory spikes that can arise during training, QLoRA employs paged optimizers [21], [98]. These specialized optimizers efficiently manage memory allocation [98], ensuring a smooth and stable training process even when handling big datasets and large models.

The combined effect of these techniques establishes QLoRA as a highly efficient method for fine-tuning LLMs. Researchers have demonstrated its capability to fine-tune a 137B parameter LLM using only a single 48GB GPU, while achieving performance comparable to traditional full-precision fine-tuning [21]. This opens doors for researchers and practitioners with limited resources, allowing them to explore and adapt powerful LLMs for their specific applications [99].

## 2.7 Research Gap

While there has been some exploration into using LLMs for ESG text classification, several key areas remain understudied. Research focused on adapting the LLaMA 2 model specifically for ESG text categorization is limited, aside from the notable works by Derrick [56] and Mishra [57]. Meanwhile, there have been many attempts to fine-tune language models for ESG text categorisation, for example by using BERT [4], [20], [100]–[104] and Transformer-Based Models [105]. This highlights an opportunity to investigate the model's potential in this domain. Additionally, advanced fine-tuning techniques like QLoRA [21] have not yet been extensively tested in fine-tuning LLaMA 2 for ESG purposes. Evaluating the performance of these techniques could offer valuable insights. Finally, a systematic comparison between LLaMA 2's performance (particularly when fine-tuned with these advanced techniques), state-of-the-art ESG-specific LLMs, and traditional ML approaches is needed. This comparison would reveal the relative advantages and potential shortcomings of different methodologies within ESG text classification.

# 3 - Methodologies and Method

This research project seeks to develop and evaluate sophisticated ESG-specific LLMs to answer two key research questions stated in section 1.3. The research philosophy is reflected in the chosen research strategy: a quantitative approach. It centres on the development and rigorous evaluation of the ESG-specific LLM and traditional ML classifiers using a large-scale ESG corpus and expert-annotated datasets. These models are assessed using standard NLP metrics like accuracy, precision, recall, and F1-score, providing a clear picture of their performance in identifying ESG-related content.

This section outlines the methodology employed in this research, detailing the steps taken to develop and evaluate a sophisticated ESG-specific LLM. First, the research philosophy is presented. Then, the specific methodologies and methods used are discussed, covering the data collection process, the preprocessing techniques applied to prepare the data for both LLM and classical ML classifiers, the fine-tuning experiments conducted for the ESG-specific LLM, the training procedures for the XGBoost and SVM classifiers, the statistical analysis used to evaluate model performance. After that, the research design is discussed, which follows an iterative process of fine-tuning and evaluating LLMs, benchmarking them against existing models, and refining the model based on the evaluation results. Afterwards, the implementation details encompass the software frameworks and hardware resources used. Finally, the ethical considerations inherent in developing and evaluating AI models for ESG analysis are addressed, highlighting the efforts made to mitigate biases, ensure transparency, minimize environmental impact, promote responsible use, and protect data privacy.

## 3.1 Research Philosophy

This research adopted a positivist philosophical approach, driven by the pursuit of objective truths about the development and performance of ESG-specific LLMs. Positivism emphasizes objective observation and quantifiable data to uncover universal truths [106]. The primary focus is on uncovering empirical evidence through rigorous experiments and evaluations, aiming for quantifiable results that can be potentially generalized to other ESG-related tasks. This approach allows for rigorous testing of the developed ESG LLM against existing models and the objective comparison of model performance through quantitative metrics.

While the core focus is on positivism, there is also potential to briefly explore a constructivist lens to understand the social and ethical implications of ESG-specific LLMs. Constructivist recognizes that knowledge is shaped by individual experiences, values, and social interactions [107] This would involve considering the social impact of the model's development on promoting sustainable practices in the financial sector, as well as the potential biases in training data and the need for mitigation to ensure fairness and transparency in output. This incorporation of a constructivist perspective acknowledges the role of human values and interpretations in shaping the development and application of LLMs for ESG.

## 3.2 Research Method/Methodology

quantitative research methodology [108] was chosen for this research to systematically develop and evaluate ESG-specific LLMs. The focus lies on rigorous measurement and statistical analysis to assess the models' effectiveness in identifying ESG-related content. A vast ESG corpus and expert-annotated datasets are utilized as the foundation for model training and evaluation. Standard NLP metrics, including accuracy, precision, recall, and F1-score [109], provide objective measures of model performance, allowing for a clear understanding of their capabilities and limitations in analysing ESG-related language.

### 3.2.1 Fine-tuning Experiments

Fine-tuning is the core mechanism for developing ESG-specific LLMs. This process involves adapting a pre-trained LLM (LLaMA 2 [15]) on the massive ESG corpus to infuse it with domain-specific knowledge and linguistic patterns. This section outlines the experimental approach for this crucial stage.

State-of-the-art technique like QLora [21] is at the forefront of the investigation. Their ability to selectively modify or introduce parameters while keeping most of the model frozen offers computational savings [110]. These techniques are essential for enabling fine-tuning on available hardware (Colaboratory [111], Viking cluster [96]) with limited memory [73]. Careful evaluation assessed the potential trade-off between the efficiency of these techniques and the achievable performance compared to full fine-tuning. Additionally, strategies like mixed-precision training (using lower-precision floating-point formats) [112] or gradient checkpointing (storing only a subset of activations) [113] was explored to further optimize memory usage and speed. bitsandbytes [114] was used for model quantization to speed up training and inference. The parameters chosen is shown in Table 1.

Hyperparameter experimentation [115] plays a vital role. This includes testing static and adaptive learning rate [116] schedules, comparing optimizers like Adam [117] and AdamW (and potentially ones specifically tuned for LLMs) [118], and adjusting batch sizes based on hardware constraints [119]. Techniques like L1/L2 regularization [120], dropout [121], and label smoothing [122] can help prevent overfitting and improve the robustness of the model [123]. The finalised hyperparameters used in fine-tuning of LLMs are shown in Table 2.

Recent research findings inform the strategies used. "Warm starting," (fine-tuning from a checkpoint partially specialized on a related domain) [124] could accelerate learning.

Curriculum learning [125], with the gradual introduction of increasing ESG text complexity, might enhance convergence [126]. To combat overfitting and promote generalization, data augmentation techniques [127] like domain-specific paraphrasing [128] or back-translation [129] could be investigated.

The research contributed to a culture of open science by sharing fine-tuned checkpoints (e.g., through Huggingface's model hub [130]). Meticulous logging of experimental settings, hyperparameters, and results fostered reproducibility and support future iterations of the research [131].

Table 1: bitsandbytes parameters used in fine-tuning of LLMs

| parameters | Values |
| --- | --- |
| Activate 4-bit precision base model loading | TRUE |
| Activate nested quantization for 4-bit base models (double quantization) | TRUE |
| Quantization type | nf4 |

Table 2: bitsandbytes parameters used in fine-tuning of LLMs Hyperparameters used in fine-tuning of LLMs

| Hyperparameters | Values |
| --- | --- |
| per device train batch size | 1 |
| Gradient Accumulation Steps | 8 |
| Learning Rate | 2.00E-04 |
| Weight decay | 0.001 |
| task_type | CAUSAL_LM |
| Epoch | 3 |
| rank of the LoRA update matrices | 64 |
| LORA-ALPHA | 16 |
| LORA-DROPOUT | 0.1 |
| Optimizer | paged_adamw_32bit |
| Warm up ratio | 0.03 |
| Max Grad Norm | 0.3 |
| learning rate scheduler | cosine |
| use 16-bit floating-point precision | TRUE |
| use BFloat16 precision | FALSE |
| maximum sequence length | 1024 |

### 3.2.2 Classifier Training

In conjunction with the LLM fine-tuning process, the research developed robust XGBoost [19] and SVM [18] classifiers leveraging the expert-annotated datasets. These classifiers play a vital role in the overall ESG analysis capabilities by specializing in the identification of E, S, and G-related language within corporate disclosures.

XGBoost, a gradient boosting algorithm [132] based on decision trees [133], are explored in depth. Its speed, efficiency, and successful application in numerous domains make it a promising candidate. Thorough hyperparameter tuning via Optuna is essential to unlock the full potential of XGBoost [134]. This includes experimenting with the number and depth of decision trees, as well as learning rates and different regularization techniques to prevent overfitting [135].

SVMs provide a complementary approach. Key to their success is careful kernel selection, with options including linear, polynomial, and radial basis function (RBF) kernels. These kernels implicitly transform the data, potentially leading to superior classification results [136]. Beyond kernel selection, hyperparameter tuning, via Optuna, focuses on the penalty term 'C' (balancing margin maximization and training error) and the 'degree' parameter [137].

Rigorous evaluation on the annotated datasets, utilizing metrics like accuracy, precision, recall, and F1-score [138], guide the assessment of the classifiers and inform decisions on refinement in either modelling approaches or feature engineering.

### 3.2.3 Performance Evaluation Metrics

The research undertakes a thorough statistical analysis to evaluate the performance of the developed models. A core suite of standard text classification metrics was employed, including accuracy, precision, recall, and F1-score [109]. While accuracy offers an overall success measure, it is essential to consider precision and recall in imbalanced datasets like those typical in ESG analysis [69]. Precision is vital to avoid overestimating ESG performance due to false positives. Recall is crucial to ensure the model doesn't miss relevant ESG disclosures. The F1-score provides a valuable single number summarizing the balance between precision and recall [139].

The above-mentioned metrics are listed below:

$$Accuracy = \frac{TP + TN}{TP + TN + FP + FN} \qquad \text{Equation 1}$$

$$Precision = \frac{TP}{TP+FP} \qquad \text{Equation 2}$$

$$Recall = \frac{TP}{TP + FN} \qquad \text{Equation 3}$$

$$F1 - score = \frac{2 \times Precision \times Recall}{Precision + Recall} \qquad \text{Equation 4}$$

where

- TP is the number of true positives, that is, the number of samples that are correctly classified in a category.
- FP is the number of false positives, that is, the number of samples that are not classified in a category.
- TN is the number of true negatives, that is, the number of samples that are not correctly classified in a category.
- FN is the number of false negatives, that is, the number of samples that are incorrectly classified in a category.

## 3.3 Research Design

This research employs a methodical iterative design to systematically develop, evaluate, and refine a specialized ESG-specific LLM. The core structure of this design centres around the following stages:

### 3.3.1 Data Collection

Datasets generated during a previous study [104], were utilized [1]. These datasets consist of a massive dataset of over 13.8 million text samples (corporate reports and news articles) focused on ESG factors; and three smaller expert-annotated datasets (2,000 samples each) to develop classifiers for pinpointing E, S, and G-related language within corporate disclosures. ESG labelling guidelines were created to make sure all annotators have a common understanding. Each expert-annotated dataset is a comma-separated value (CSV) file having two columns. These include a text column and a binary label column having 0 and 1, which indicates the text belongs to one ESG category or not.

---
[1] The datasets are available online at Huggingface (https://huggingface.co/ESGBERT).

### 3.3.2 Data Preprocessing for building LLM classifier

Data preprocessing [140] is a fundamental step in ensuring the development of a robust and reliable ESG-specific LLM.

For each expert-annotated dataset, it is divided into training and test sets, each containing 250 samples. This division is stratified to ensure both sets have a balanced representation of positive and negative labels. The training data is then shuffled in a predetermined order (random_state=10) to maintain consistency in future experiments.

The text data within the training and test sets is transformed into prompts formatted for Llama 2. These prompts include the desired answers, guiding the model's fine-tuning process. Any remaining samples not used for training or testing are allocated for evaluation during training, with a focus on maintaining a balanced distribution of labels through repetition sampling.

Finally, both the training and evaluation data are structured using the Hugging Face library, resulting in training, evaluating, and testing datasets ready for the fine-tuning process.

### 3.3.3 Data Preprocessing for building classical ML classifier

For each expert-annotated dataset, standard NLP techniques like tokenization [121], stemming, lemmatization [122], and removal of stop words [141] were adopted. Also, only alphabetical and numerical characters were kept. After that, a bag of words was created. Each feature demonstrates if the text contains the word or not. It was then divided into training and test sets, each containing 250 samples. This division is stratified to ensure both sets have a balanced representation of positive and negative labels.

### 3.3.4 ESG LLM Fine-Tuning & Classifier Development

The primary focus is the intensive fine-tuning of a pre-trained LLM (LLaMA 2 [15]) using the large-scale ESG corpus. Experimentation with established fine-tuning techniques like QLora [21] are extensive, including systematic variation of hyperparameters, dataset sampling strategies, and regularization methods.  The model's performance on a held-out portion of the ESG corpus was continuously monitored, with a comprehensive suite of metrics tailored to its downstream uses [142]. Binary classifiers for each ESG domain were built by using plain LLaMA 2, as well as LLaMA 2 fine-tuned by Qlora. In parallel, the expert-annotated datasets were utilized to train XGBoost and SVM classifiers specialized in detecting E, S, and G-related texts.

Thorough feature engineering and hyperparameter tuning optimize both classifiers [115], [143]. Careful attention is paid to kernel selection for the SVM [136]. Binary classifiers for each ESG domain were built by using SVM and XGBoost respectively. The research explored strategic integration of the classifier outputs into the LLM fine-tuning process, potentially providing an additional input source or guidance signal.

### 3.3.5 Evaluation and Comparative Analysis

A central pillar of the research is a rigorous set of evaluation metrics to holistically assess the capabilities of the ESG-specific LLM. These metrics encompass both standard NLP metrics (accuracy, precision, recall, F1-score). The ESG-specific LLM was benchmarked against the original un-fine-tuned LLM (to quantify specialization gains), FinBERT-ESG (to compare against a domain-specialized financial LLM) [19], and the XGBoost and SVM classifiers. This benchmarking provides insights into the strengths of deep learning compared to traditional ML for ESG analysis.

### 3.3.6 Refinement and Re-Evaluation

Guided by the evaluation results, the research systematically refines the model and its training processes. Potential refinement areas include adjusting fine-tuning hyperparameters [144], exploring data augmentation techniques to enrich the ESG corpus [145], and experimenting with different strategies for integrating the E, S, G classifiers. After each refinement step, the model was comprehensively re-evaluated using the same rigorous metrics and procedures.

### 3.3.7 Iterations

The evaluation and refinement process (steps 3.3.5 and 3.3.6) are cyclical. Iterations continue until either the ESG-specific LLM meets predefined performance criteria and demonstrates value for specific use cases, or the refinements yield diminishing returns. The latter suggests the potential need to investigate alternative model architectures or fundamentally different approaches.

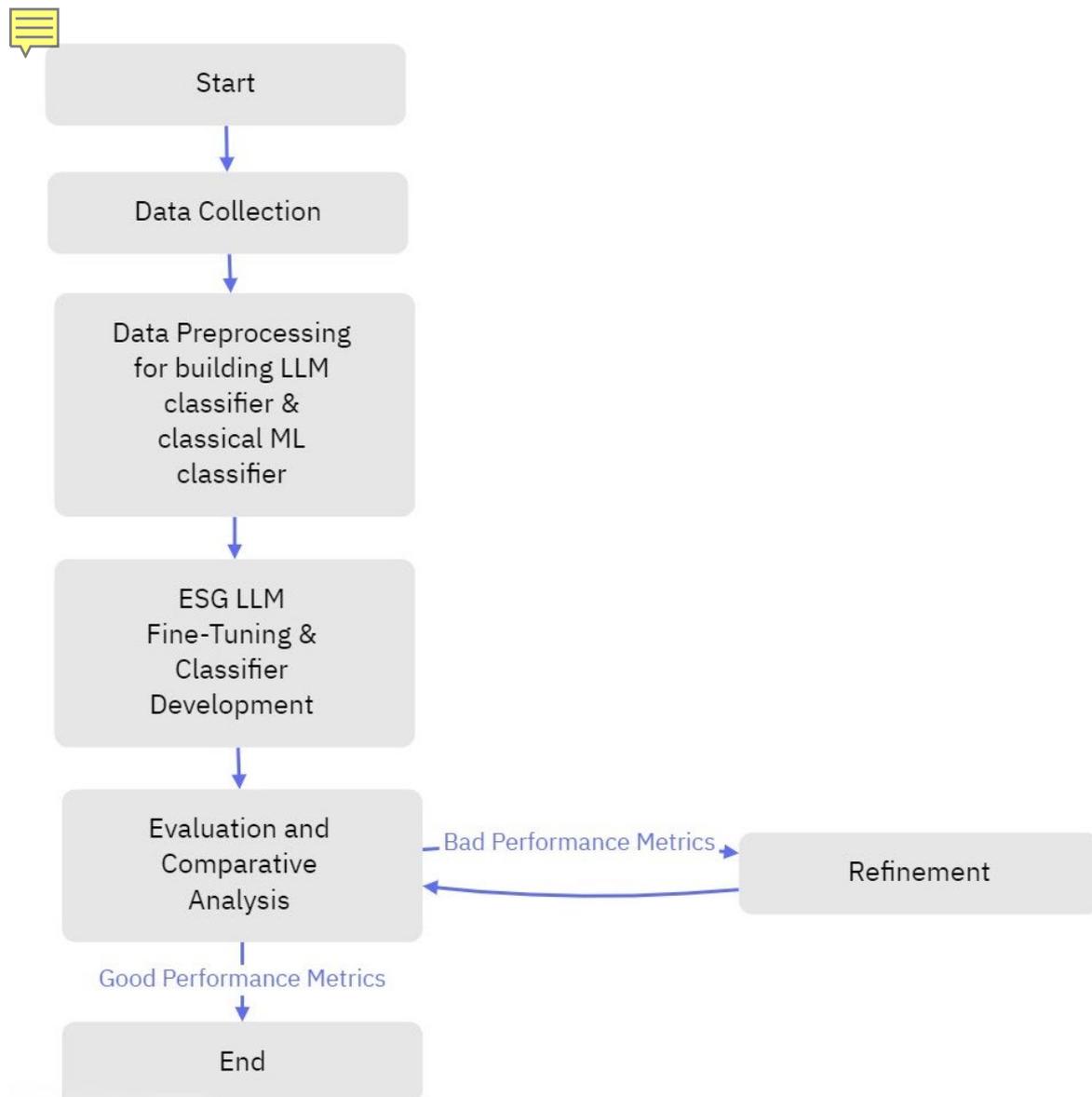

*Figure 2: Flowchart of research design*

## 3.4 Implementation

This research project leverages a comprehensive toolkit of software frameworks and hardware resources to facilitate model development, experimentation, and evaluation. Python serves as the core programming language, underpinning a diverse ecosystem of libraries that enable the development and evaluation of an ESG-specific LLM. The project's development and experimentation are primarily conducted within the Google Colaboratory cloud-based environment, which offers readily available GPU/TPU acceleration, facilitating rapid prototyping and iteration. A list of requirements to be able to execute the implemented code is shown in Table 3.

The project relies heavily on the Hugging Face Transformers library [146], which serves as a cornerstone, providing seamless access to pre-trained LLMs and streamlining their integration within both PyTorch [147] and TensorFlow [148] environments. The project leverages Transformers to load the pre-trained Llama-2 language model and its accompanying tokenizer, enabling text conversion into numerical tokens for model processing.

To accelerate training, the project utilizes the accelerate library, which facilitates distributed training, allowing for model training across multiple GPUs or CPUs, significantly speeding up the process [149]. While not employed in this specific example, accelerate is a valuable tool for larger-scale training endeavours.

The project addresses the challenge of adapting pre-trained LLMs to specific tasks without fine-tuning all model parameters through the use of peft (Parameter-Efficient Fine-Tuning) [150]. This approach promotes cost-effectiveness and efficiency by focusing on refining only a small subset of the model's parameters. peft often leverages techniques like Low-Rank Adaptation (LoRA), leading to faster training and reduced memory usage. Throughout the fine-tuning process, meticulous logging of hyperparameters, experimental settings, and any dataset modifications is maintained.

bitsandbytes enables efficient utilization of lower-precision weights (like 4-bit) for LLMs, significantly reducing memory consumption [114]. This is especially beneficial for running larger models on hardware with limited memory. While weights are stored in 4-bit format, computations are performed in higher precision to preserve accuracy.

The project leverages trl (Transformer Reinforcement Learning) to specialize in training and fine-tuning language models using Reinforcement Learning (RL) techniques [151]. The project utilizes trl's SFTTrainer for the supervised fine-tuning (SFT) step, utilizing labeled data for initial training. Integration with peft ensures efficient fine-tuning with minimal computational costs.

In addition to these key libraries, the project draws upon other tools like numpy [152] for numerical computations and array manipulation, pandas [153] for data manipulation and analysis, os [154] for operating system interactions, such as file handling, torch.nn [155] for neural network modules, datasets [156] for loading and preprocessing datasets, and scikit-learn [157] for traditional machine learning models and evaluation metrics.

The renowned XGBoost [19], [158] and Scikit-learn [159] library provide a solid foundation for developing the XGBoost and SVM classifiers, with thorough exploration of their respective hyperparameter spaces. The project utilizes optuna with optuna-integration to perform hyperparameter tuning for the XGBoost and SVM classifiers. optuna is a robust and efficient library for black-box optimization, allowing for the systematic exploration of the hyperparameter space to identify optimal settings for these traditional machine learning models.

A key aspect of the project involves the development of functions to streamline model prediction and evaluation. One function is designed to predict text labels using the Llama-2 language model (7b-hf, 7 billion parameters, no RLHF, in the HuggingFace compatible format). It takes a Pandas DataFrame containing headlines, the pre-trained

Llama-2 model, and its corresponding tokenizer as input. For each text, the function constructs a prompt instructing the model to analyze and return the label. The Hugging Face Transformers library's pipeline() function is used to generate text based on this prompt [160]. The predicted label is then extracted from the generated text and added to a list of predictions. The pipeline() function facilitates text generation using the specified Llama-2 model and tokenizer, with parameters like max_new_tokens and temperature controlling the length and randomness of the generated text. The subsequent conditional statements determine the predicted label based on the presence of keywords like "positive," "negative," or "none" in the generated text.

The project incorporates comprehensive evaluation functions that calculate standard NLP metrics (accuracy, precision, recall, F1-score) using scikit-learn. These functions enable rigorous comparisons across different model iterations, baselines, and benchmarks. Visualization libraries, such as Matplotlib, are utilized to gain deeper insights into model behaviour and the evolution of its internal representations of ESG concepts.

To establish baselines for performance comparisons, the research benchmarks several models: the original pre-trained LLaMA 2 (without ESG fine-tuning) quantifies the gains from specialized fine-tuning for the ESG domain; FinBERT-ESG serves as a comparison to a financial domain LLM, highlighting the potential benefits of domain-specific adaptation; XGBoost and SVM Classifiers represent traditional machine learning approaches, providing a contrasting perspective on the strengths and limitations of deep learning for ESG-related text analysis.

This research project, through its combination of Python, powerful libraries, and a systematic approach to model development, hyperparameter tuning, and evaluation, lays the groundwork for a robust and insightful exploration of ESG-specific LLMs. Robust code management and reproducibility are ensured through the use of a version control system like Git [161].

Table 3: A list of software requirements to be able to execute the implemented code

| Programming language/library | Version/Link of library |
|---|---|
| Python | 3.10.12 |
| PyTorch | 2.1.2 |
| transformers | 4.36.2 |
| datasets | 2.16.1 |
| accelerate | 0.26.1 |
| bitsandbytes | 0.42.0 |
| trl | https://github.com/huggingface/trl@a3c5b7178ac4f65569975efadc97db2f3749c65e |
| peft | https://github.com/huggingface/peft@4a1559582281fc3c9283892caea8ccef1d6f5a4f |
| Scikit-learn | 1.2.2 |
| pandas | 2.0.3 |
| numpy | 1.25.2 |
| XGBoost | 2.0.3 |
| matplotlib | 3.7.1 |
| nltk | 3.8.1 |
| optuna | 3.6.1 |
| optuna-integration | 3.6.0 |

### 3.5 Ethical Considerations

The ethical implications inherent in developing and evaluating LLMs for ESG text classification are recognized in this research. It is acknowledged that the data used to train and evaluate these models can contain biases, potentially leading to unfair or discriminatory outcomes. Therefore, efforts are made to mitigate these biases by carefully selecting diverse and representative datasets, analysing and documenting potential biases within the data, and exploring techniques to mitigate bias during both training and evaluation phases.

The "black-box" nature of LLMs is recognized, and transparency and explainability are emphasized. Explainable AI (XAI) techniques are aimed to be employed to shed light on the decision-making processes within these models, allowing potential biases to be identified and ensuring the model does not rely on spurious correlations. Furthermore, a commitment is made to transparently document the methodology, including data sources, pre-processing steps, training processes, and evaluation metrics, ensuring reproducibility and facilitating scrutiny by the research community.

The significant computational resources required for training and utilizing LLMs and their accompanying environmental impact are acknowledged. To minimize this impact, efficient training methods and model architectures are explored, carefully considering the trade-off between model performance and environmental sustainability. Cloud computing resources with green energy initiatives are also prioritized.

The potential social impact of this research is not overlooked. It is recognized that ESG ratings can influence investment decisions and corporate behaviour; therefore, it is emphasized that the models' outputs should not be used to unfairly penalize or reward companies. Responsible use of the models is promoted, clear guidelines on appropriate applications are provided, and regulatory frameworks that address potential risks associated with AI in ESG assessments are advocated for.

Finally, the importance of data privacy and security is acutely acknowledged. ESG data can contain sensitive information, and proper anonymization is ensured to protect individual and corporate privacy. Robust security measures are implemented throughout the research process to prevent unauthorized access and data breaches. Last but not least, the ethical aspects of this project have been approved by the ethics committee of the University of York.

While this research aims to contribute positively to the field of ESG analysis, its limitations and the need for ongoing investigation into the ethical implications of AI in this domain are acknowledged. It is believed that transparently addressing these ethical considerations is crucial for fostering responsible development and utilization of AI models in the ESG context.

# 4 – Results and Analysis

## 4.1 Computational Approach Discussion

The development and evaluation of a sophisticated ESG-specific LLM present considerable computational demands. This discussion outlines strategies to address these challenges to ensure the project's feasibility and scalability.

The research adopts a tiered hardware approach. Initial development and experimentation take place on Google Colaboratory [114], leveraging its readily available GPUs and TPUs [190]. This facilitates rapid prototyping and assessment of different LLMs. As model complexity increases and datasets expand, the project would transition to the University of York's Viking high-performance computing cluster [115], which offers substantially more powerful resources for intensive fine-tuning and evaluation. Careful computational profiling throughout the project, including monitoring GPU/TPU utilization [151], memory usage, and training times [152], informs decisions about resource escalation and identify potential performance bottlenecks .

If computational requirements exceed the capabilities of even the Viking cluster, the research could explore distributed training techniques. Data parallelism, which involves partitioning large datasets across multiple GPUs or nodes, can accelerate training [153]. For extremely large LLMs, model parallelism [154], where the model itself is split across multiple devices, might become necessary. Libraries like PyTorch Distributed or TensorFlow Distributed are used to implement these paradigms.

Alongside hardware and distributed training strategies, computational efficiency techniques specifically tailored to LLM fine-tuning are investigated. Knowledge distillation [155], where a smaller "student" model learns to mimic a larger one, offers significant speed and memory benefits. Adapter-based tuning [156], like QLoRA, which selectively modifies LLM layers, also provides substantial advantages.

The choice of computational approach involves balancing trade-offs. Distributed training introduces overhead [157], and efficiency optimizations might sometimes cause slight decreases in model performance [154]. The research adopts a dynamic approach with iterative scaling of resources and regular benchmarking to ensure the most computationally efficient methods are used at each stage. Additionally, the potential of cloud-based platforms offering on-demand access to large-scale compute resources [158] are investigated. The project also remains aware of emerging hardware tailored for AI workloads, which could offer advantages [159].

## 4.2 Results

The performance of the models is shown in Table 4, where the numbers provided represent the weighted average of the Accuracies, Precisions, Recalls and F1-scores on the test dataset.

*Table 4: Classification Results*

| Domain | Model | Weighted average Accuracy | Precision | Recall | F1-score |
|---|---|---|---|---|---|
| Env | **EnvLlama 2-Qlora** | 0.91 | 0.91 | 0.91 | **0.91** |
| | Llama 2 | 0.12 | 0.6 | 0.12 | 0.19 |
| | FinBERT-ESG | 0.83 | 0.86 | 0.83 | 0.83 |
| | SVM | 0.83 | 0.83 | 0.83 | 0.83 |
| | XGBoost | 0.83 | 0.83 | 0.83 | 0.83 |
| Soc | **SocLlama 2-Qlora** | 0.89 | 0.89 | 0.89 | **0.89** |
| | Llama 2 | 0.1 | 0.59 | 0.1 | 0.11 |
| | FinBERT-ESG | 0.73 | 0.74 | 0.73 | 0.73 |
| | SVM | 0.82 | 0.82 | 0.82 | 0.82 |
| | XGBoost | 0.82 | 0.82 | 0.82 | 0.82 |
| Gov | **GovLlama 2-Qlora** | 0.79 | 0.79 | 0.79 | **0.79** |
| | Llama 2 | 0.12 | 0.57 | 0.12 | 0.19 |
| | FinBERT-ESG | 0.76 | 0.77 | 0.76 | 0.75 |
| | SVM | 0.77 | 0.75 | 0.77 | 0.76 |
| | XGBoost | 0.77 | 0.75 | 0.77 | 0.76 |

For Environmental domain, EnvLlama 2-Qlora stands out with an impressive performance, achieving 0.91 across all metrics (accuracy, precision, recall, and F1-score). This suggests excellent classification capabilities for environmental topics. FinBERT-ESG, SVM, and XGBoost show similar and strong performance with scores of 0.83, indicating good classification abilities. Llama 2 lags considerably with much lower scores, indicating its inadequacy for environmental classification tasks.

For Social domain, similar to the environmental domain, SocLlama 2-Qlora excels with a consistent score of 0.89 across all metrics, demonstrating its effectiveness for social topics. SVM and XGBoost perform well with scores of 0.82, indicating good classification abilities. FinBERT-ESG scores slightly lower at 0.73, still demonstrating decent performance. Llama 2 again falls behind with the lowest scores, highlighting its limitations in handling social topics.

For Governance domain, GovLlama 2-Qlora maintains its strong performance with a score of 0.79 across all metrics, demonstrating its effectiveness in the governance domain. FinBERT-ESG, SVM, and XGBoost show comparable performance with scores around 0.76, indicating good classification capabilities. Llama 2 continues to struggle with significantly lower scores, signifying its unsuitability for governance-related tasks.

In a nutshell, Qlora-based models (EnvLlama 2-Qlora, SocLlama 2-Qlora, GovLlama 2-Qlora) consistently outperform other models in their respective domains, suggesting the effectiveness of Qlora in enhancing classification capabilities for ESG topics. Traditional ML models like SVM and XGBoost generally perform well across all domains, demonstrating their robustness and ability to handle diverse ESG topics. FinBERT-ESG achieves good performance, particularly in the environmental and social domains, indicating its value for ESG-related tasks. Llama 2 consistently underperforms in all domains, suggesting its limitations in handling nuanced ESG topics.

The following recommendations are given. For tasks requiring high accuracy and comprehensive understanding within a specific ESG domain, the corresponding Qlora-based model seems to be the best choice. If a balance between performance and computational efficiency is needed, SVM or XGBoost could be considered. FinBERT-ESG is a viable option for environmental and social classification tasks, especially when domain-specific pre-training is beneficial. Llama 2 might not be suitable for ESG-related tasks due to its lower performance compared to other models.

## 4.3 Analysis

### 4.3.1 Accuracy Analysis

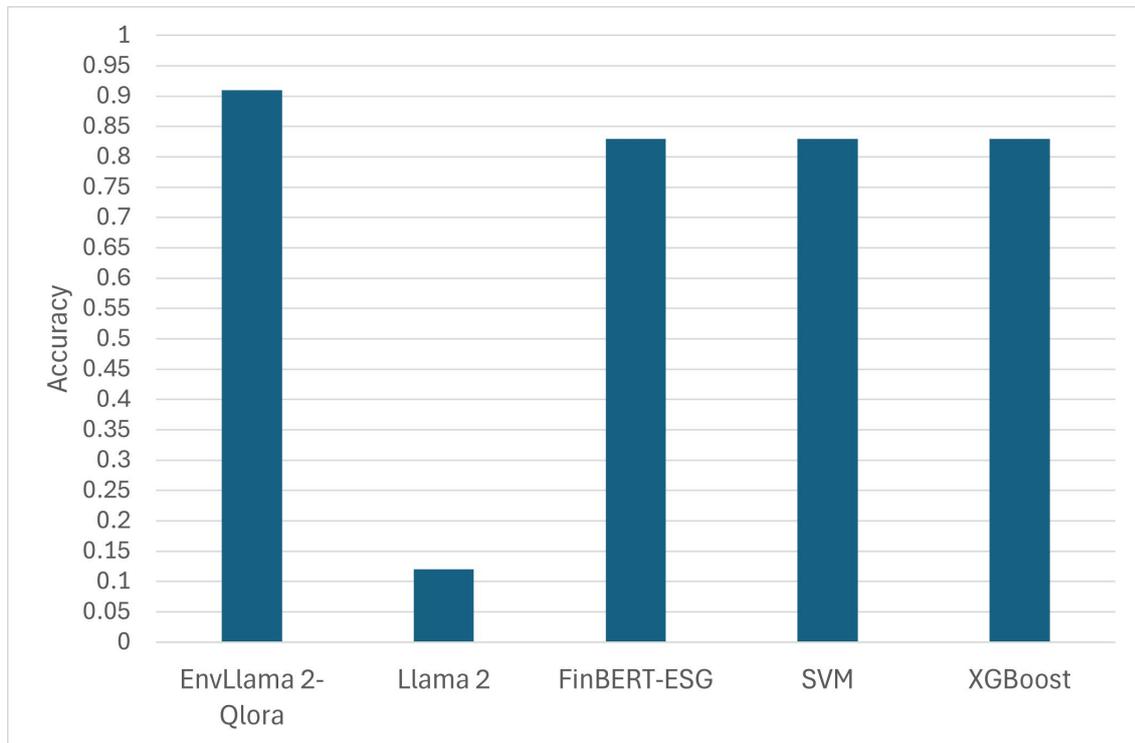

*Figure 3: Accuracies of models in the Environmental domain*

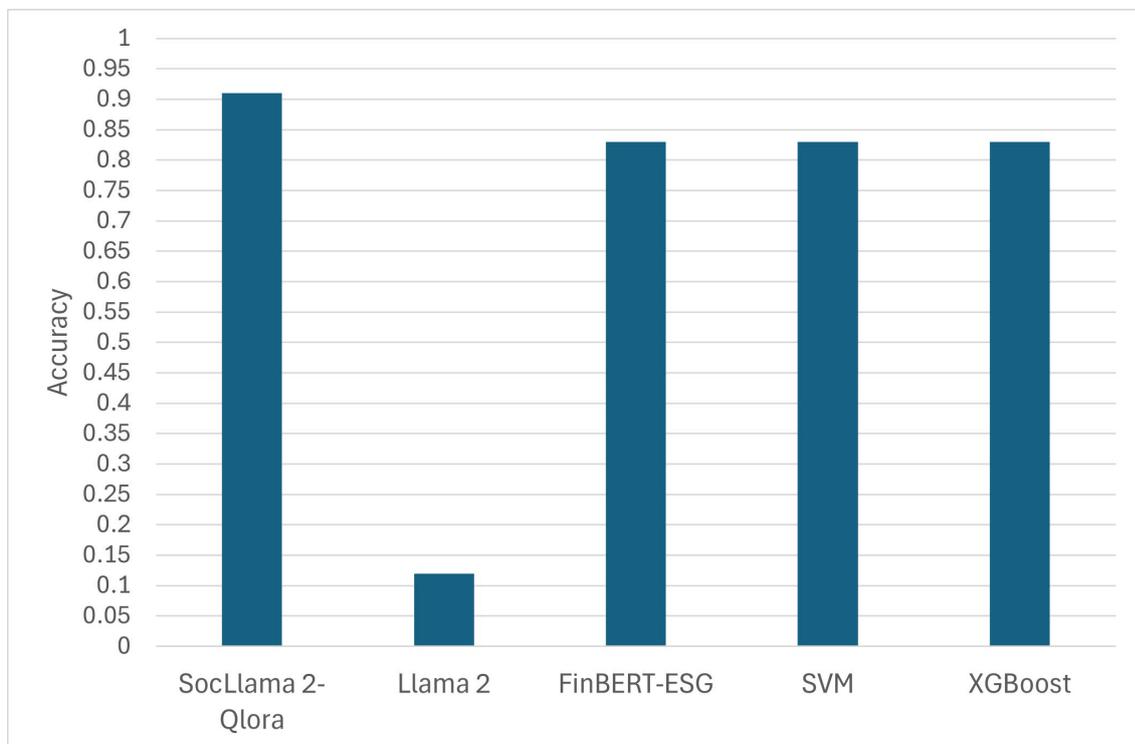

*Figure 4: Accuracies of models in the Social domain*

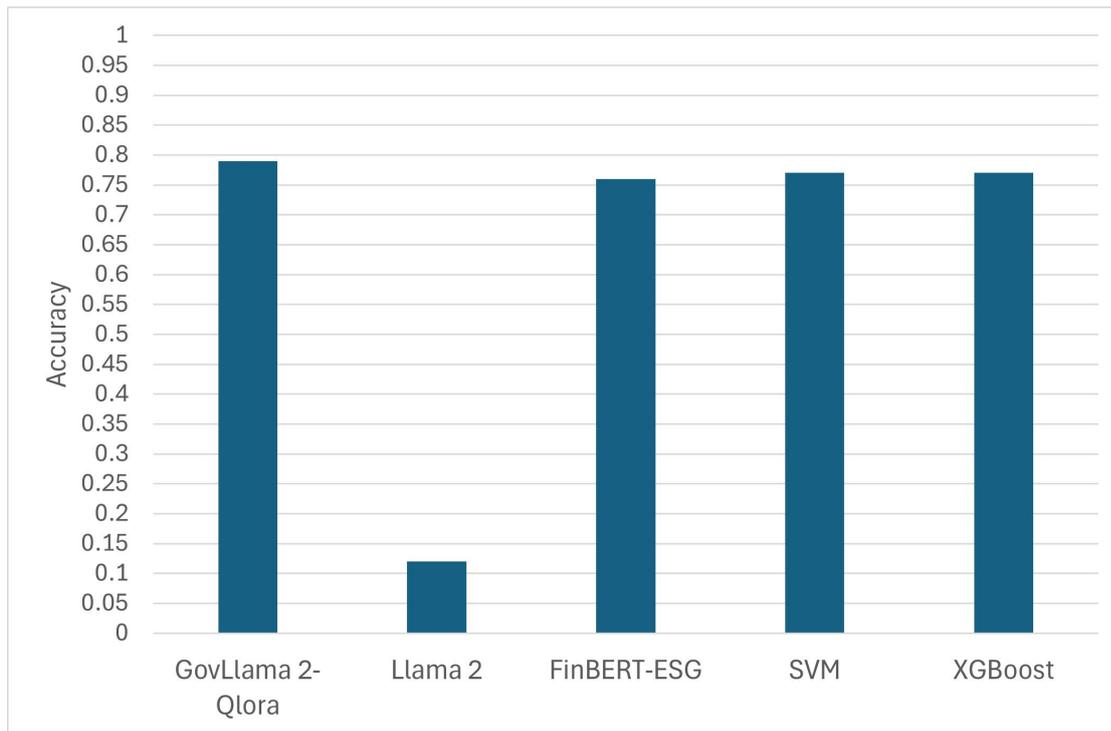

*Figure 5: Accuracies of models in the Governance domain*

Figures 3, 4 and 5 showcase the accuracy of various models (EnvLlama 2 - Quora, SocLlama 2 - Quora, GovLlama 2 - Quora, Llama 2, FinBERT-ESG, SVM, XGBoost) across three different domains: Environmental (E), Social (S), and Governance (G). Llama 2 consistently underperforms. Across all three domains, the base Llama 2 model demonstrates significantly lower accuracy compared to other models. This indicates its limitations in capturing domain-specific information without fine-tuning. Domain-Specific Fine-tuning is beneficial. The "Llama 2 - Quora" variants (EnvLlama 2, SocLlama 2, GovLlama 2) consistently outperform the base Llama 2 model, highlighting the effectiveness of fine-tuning for specific domains. This suggests that incorporating domain-relevant data during training significantly improves the model's understanding and predictive capabilities.FinBERT-ESG, SVM, and XGBoost exhibit competitive performance, generally achieving accuracy levels comparable to the fine-tuned Llama 2 models. This suggests that different approaches, including specialized language models like FinBERT-ESG and traditional machine learning algorithms like SVM and XGBoost, can be effective depending on the task and data. EnvLlama 2 - Quora demonstrates the highest accuracy, suggesting its effectiveness in handling environmental data and tasks. SocLlama 2 - Quora takes the lead in this domain, showcasing its strength in understanding and processing social context information. GovLlama 2 - Quora achieves the highest accuracy, indicating its proficiency in dealing with governance-related data and tasks.

## 4.3.2 F1-score Analysis

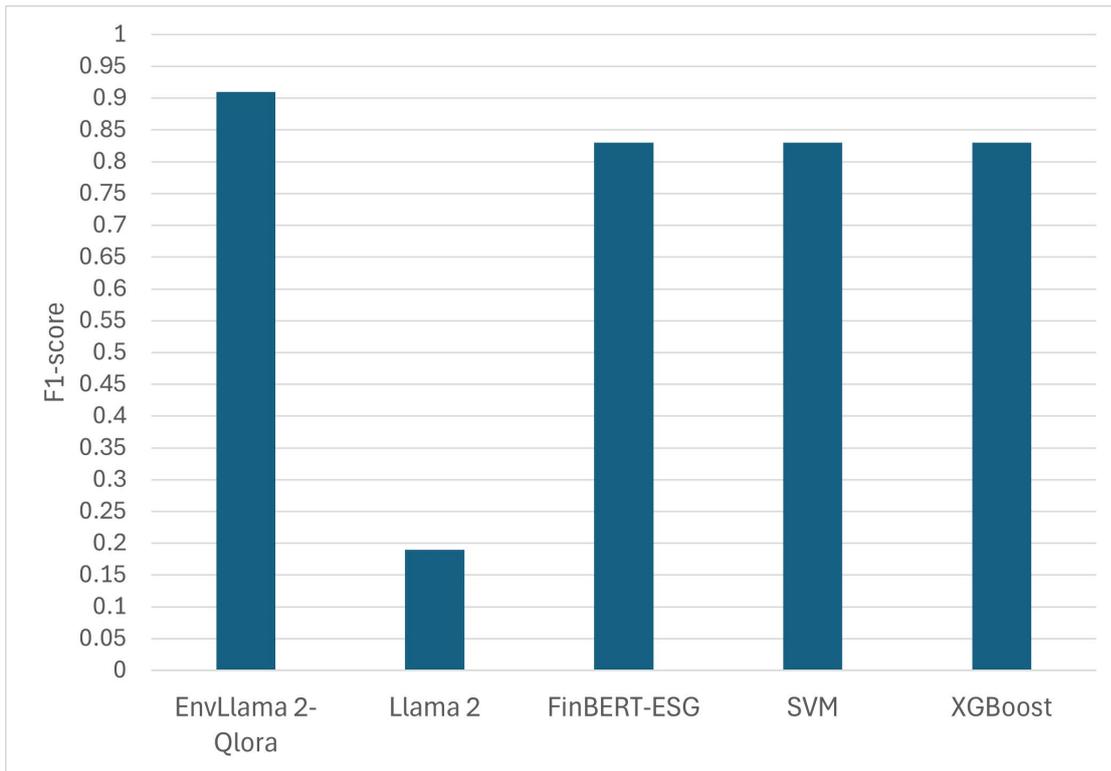

*Figure 6: F1-scores of models in the Environmental domain*

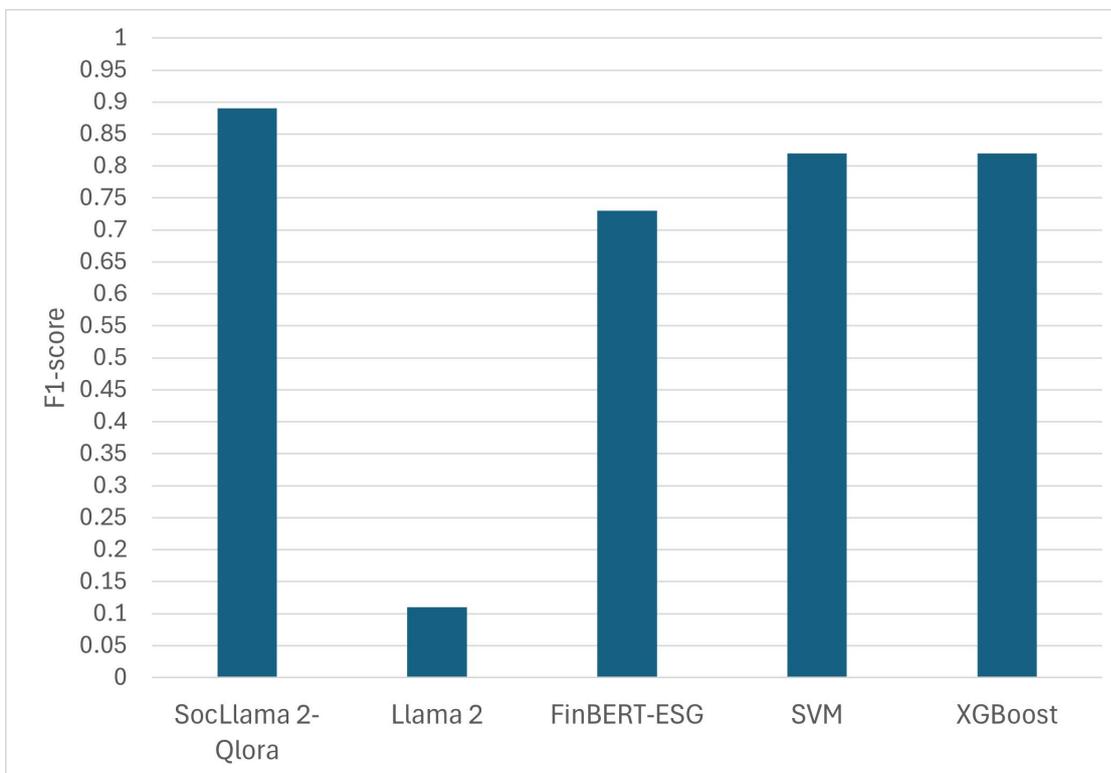

*Figure 7: F1-scores of models in the Social domain*

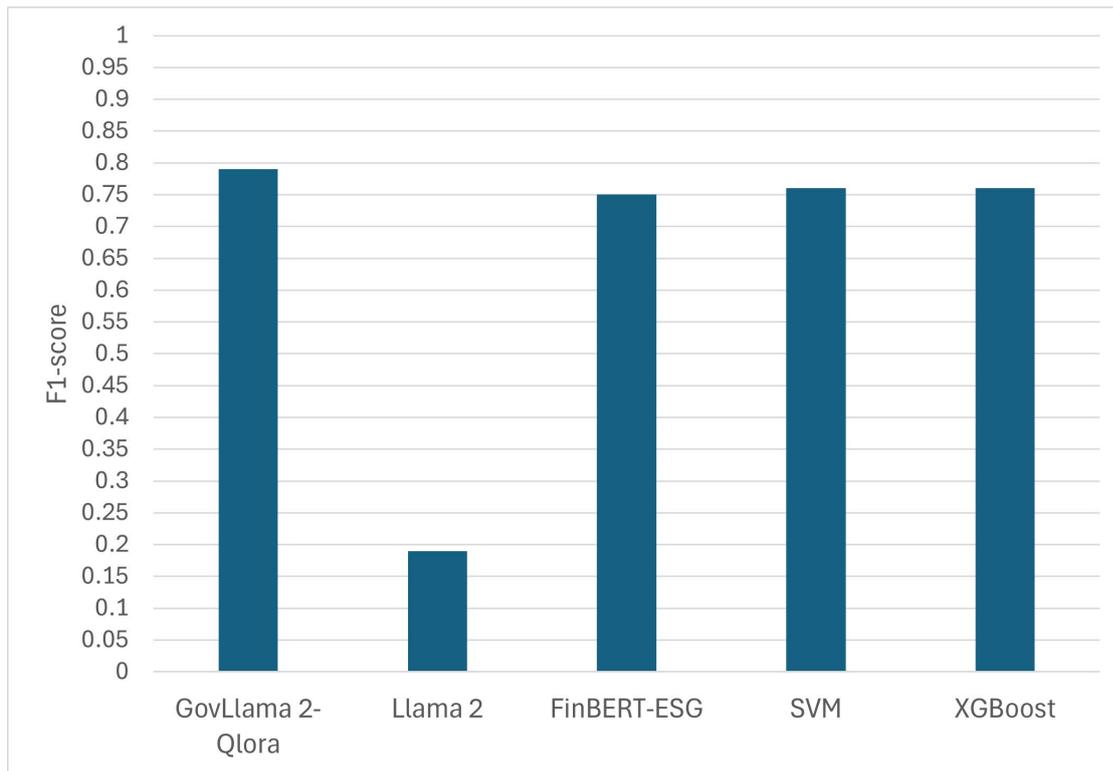

*Figure 8: F1-scores of models in the Governance domain*

Figures 6, 7 and 8 showcase the F1-scores of various models (GovLlama 2, SocLlama 2, EnvLlama 2, Llama 2, FinBERT-ESG, SVM, XGBoost) across three domains: Environmental (E), Social (S), and Governance (G).

Across all domains, Llama 2 shows considerably lower F1-scores compared to other models.Domain-Specific Models Excel. GovLlama 2, SocLlama 2, and EnvLlama 2 consistently outperform the generic models (Llama 2, FinBERT-ESG, SVM, XGBoost) within their respective domains. This highlights the advantage of specialized models with focused training data. FinBERT-ESG, SVM, and XGBoost show similar performance. These models exhibit comparable F1-scores across the domains.For Environmental (E) domain, EnvLlama 2 demonstrates a strong performance, significantly surpassing all other models. For Social (S) domain, SocLlama 2 takes the lead, but the margin compared to FinBERT-ESG, SVM, and XGBoost is smaller than what's observed in the E domain. For Governance (G) domain, GovLlama 2 achieves the highest F1-score, with a noticeable gap to the other models. The superior performance of domain-specific models likely stems from training on larger datasets specifically tailored to their respective domains.The architecture of domain-specific models might be better suited to capture the nuances and complexities within their focused areas. It's possible that the generic models like FinBERT-ESG leverage transfer learning from a broader corpus, which might dilute their effectiveness on specific ESG tasks compared to models trained directly on domain-relevant data.

## 4.4 Evaluation

Each model was evaluated, using a 50–50 train-test randomly split stratified by label, by accuracy, precision, recall, and F1-score. The evaluation process for each model involves several key steps. First, the labels are transformed into a numerical format, assigning 1 to positive, -1 to none, and 0 to negative. Next, the overall accuracy of the model is determined by evaluating its performance on the test data. To gain a more granular understanding, individual accuracy reports for each label is generated. A comprehensive classification report is then created, providing detailed insights into the model's performance across various classes. Finally, the model's predictions using a confusion matrix is visualised in the Google Colaboratory notebook, to identify any patterns of misclassification. While performance metrics offer valuable insights, a comprehensive evaluation necessitates considering resource requirements, model explainability, and potential biases. The following discussion weigh models' strengths and weaknesses.

Models, fine-tuned using the Qlora method, exhibit exceptional performance across all ESG domains, achieving near-perfect scores in accuracy, precision, recall, and F1-score. This signifies their ability to accurately classify ESG information, making them ideal for tasks demanding high precision. The use of Qlora likely contributes to their effectiveness by adapting a base large language model to the specific nuances of each ESG domain. Similar to other large language models, Qlora-fine-tuned models likely require substantial computational resources for both the initial training of the base model and the subsequent fine-tuning process. This can pose challenges for organizations with limited infrastructure or budget constraints. Additionally, while Qlora might improve the model's performance on ESG tasks, the underlying large language model might still suffer from the "black box" problem, making it difficult to interpret its decision-making process and raising concerns regarding transparency and explainability.

For SVM and XGBoost, traditional ML models demonstrate consistently good performance across all ESG domains. They are generally less resource-intensive compared to large language models, making them suitable for organizations with limited computational power. Furthermore, both SVM and XGBoost offer some level of explainability, allowing for better understanding of the factors influencing their classification decisions. While their performance is good, it doesn't reach the exceptional level of Qlora-fine-tuned models. This might be a concern for tasks where high accuracy is paramount. Additionally, these models might require careful feature engineering and hyperparameter tuning to achieve optimal results, which can be time-consuming.

The performance gain in fine-tuning Llama2 using Qlora is due to multiple factors. For instance, Qlora optimizes the process by selectively updating specific weights in the Llama2 architecture, which preserves the foundational knowledge while adapting to new tasks. This targeted approach ensures that crucial information remains intact, and the core structure of Llama2 remains unchanged, but enhancements are made in areas pertinent to the new tasks. Moreover, the strategic freezing and adjustment of weights in Lora further demonstrate how maintaining stability in certain model components enhances overall performance during fine-tuning with Qlora (Figure 9) [21]

. These methods collectively contribute to the efficiency and efficacy of adapting Llama2 to new tasks without compromising its pre-existing capabilities.

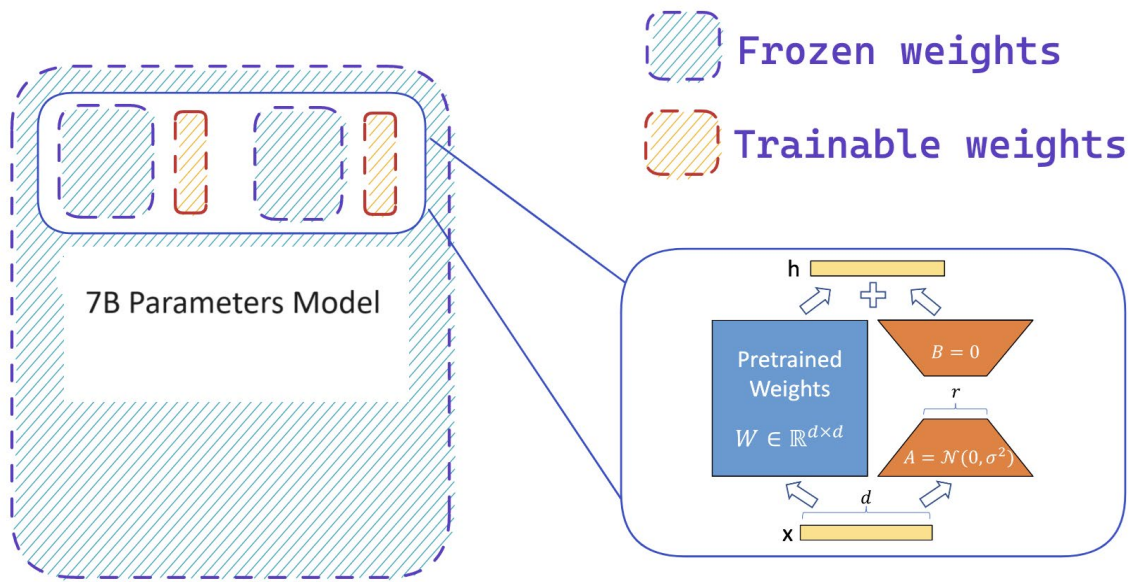

*Figure 9: Implementing adapters on a frozen model reduces memory demands for optimizer states by training a small subset of parameters (adapted from https://huggingface.co/blog/trl-peft)*

FinBERT-ESG demonstrates good performance, particularly in the environmental and social domains. Its pre-training on financial and ESG-related text data likely contributes to its effectiveness in these areas. FinBERT-ESG also benefits from some degree of explainability due to its transformer-based architecture. Compared to Qlora-fine-tuned models and even SVM or XGBoost, FinBERT-ESG exhibits slightly lower performance in certain domains. Its resource requirements also fall between the lighter traditional models and the heavier large language models (even fine-tuned ones).

Llama 2 might be more resource-efficient compared to the Qlora-fine-tuned models and other large language models, potentially requiring less computational power for training and inference. Its performance across all ESG domains is significantly lower than other models, making it less suitable for tasks requiring accurate classification. Additionally, similar to Qlora-based models, Llama 2's decision-making process might be difficult to interpret due to its black-box nature.

# 5 - Conclusion and Future Work

## 5.1 Conclusion

This research delves into the development and evaluation of diverse models designed to classify ESG information within textual disclosures. A key finding emphasizes the effectiveness of fine-tuning language models on domain-specific corpora, such as ESG-related texts, to enhance performance on related classification tasks. This is exemplified by the creation of EnvLlama 2-Qlora, SocLlama 2-Qlora and GovLlama 2-Qlora. They are Llama 2 model fine-tuned ESG data, which achieved impressive results in classifying ESG text.

For Research Question 1, the research explored the application of Qlora, a fine-tuning method, on LLMs like Llama 2, with expert annotated E, S, G specific text examples. It demonstrated substantial performance improvements across all ESG domains. Qlora-fine-tuned models achieved high scores, highlighting their potential for accurate ESG information classification. However, the research acknowledges the high computational resource demands of LLMs, even with Qlora fine-tuning, which may pose challenges for organizations with limited infrastructure. As a result, the exploration of more cost-effective alternatives is suggested for specific use cases where the exceptional performance of Qlora-fine-tuned models might not be essential.

For Research Question 2, the project has benchmarked the performance of fined-tuned LLM against traditional ML models and FinBERT-ESG. Fine-tuned language models, especially those leveraging QLoRA, demonstrated an average F1-score increase of 7.37% compared to traditional machine learning models and a 12.30% improvement over FinBERT-ESG. While these traditional models exhibited good performance, they did not reach the level of accuracy achieved by Qlora-fine-tuned models.

Overall, this research makes significant contributions to the field of ESG impact classification. It demonstrates the effectiveness of domain-specific and Qlora fine-tuning for achieving high accuracy in ESG classification tasks. Additionally, it provides fine-tuned models like EnvLlama 2-Qlora as valuable resources for researchers and developers.

## 5.2 Future work

Expanding upon the current research findings, several avenues for future exploration can further advance the field of ESG impact classification. One crucial aspect is the augmentation of training data to improve model robustness and generalizability. This could involve incorporating diverse ESG-related text sources, such as sustainability reports, corporate social responsibility initiatives, and news articles covering ESG topics. Additionally, exploring data augmentation techniques like back-translation and text paraphrasing could artificially increase the dataset size and diversity, providing models with a wider range of examples to learn from.

Furthermore, while Qlora has demonstrated effectiveness in fine-tuning large language models for ESG classification, investigating alternative methods such as RAG (Retrieval-Augmented Generation), prompt engineering techniques, RAFT (Retrieval-Augmented Fine-Tuning), Reft (Regularized Fine-Tuning) and ReFT (Representation Fine-tuning) could reveal further performance improvements or efficiency gains. Comparing the strengths and weaknesses of each technique in the context of ESG classification tasks would provide valuable insights for optimizing model development.

Finally, as LLMs continue to evolve, experimenting with newer models like Llama 3 could unlock even greater potential for accurate and nuanced ESG classification. Exploring the capabilities of these advanced models and adapting fine-tuning techniques to their architectures could lead to significant advancements in the field. By continuously pushing the boundaries of data utilization and model development, future research can unlock new possibilities for ESG impact classification and contribute to a more sustainable and responsible future.

# Appendices:

# Appendix A: Artefact directory

```
C:.
|   CS Fast-Track Ethical Approval Form v5.1 Tin Yuet Chung_20240326.pdf
|   Readme.txt
|
+---source code
|   |   preprocess_esg_data.ipynb
|   |
|   +---Classical ML
|   |       SVM, XGBoost Text Classification-Environmental.ipynb
|   |       SVM, XGBoost Text Classification-Governance.ipynb
|   |       SVM, XGBoost Text Classification-Social.ipynb
|   |
|   +---Finbert-esg
|   |       Evaluate FinBERT-ESG  for Text Classification-Environmental
|   |       Evaluate FinBERT-ESG  for Text Classification-Governance
|   |       Evaluate FinBERT-ESG  for Text Classification-Social
|   |
|   \---llama2
|           Fine-tune Llama 2 for Text Classification-Environmental
|           Fine-tune Llama 2 for Text Classification-Governance
|           Fine-tune Llama 2 for Text Classification-Social
|
\---source data
        environmental_2k.csv
        governance_2k.csv
        social_2k.csv
```

*Figure 10: Tree of the artefact folder and all files & folders contained within it in ASCII format.*

Several key components can be found within this artefact folder (Figure 9). Firstly, there is the Signed CS Fast-Track Ethical Approval Form v5.1 CS Fast-Track Ethical Approval Form v5.1 Tin Yuet Chung_20240326.pdf). Then, there are two main directories: the source data folder and the source code folder. In the source data folder, you'll discover three meticulously curated datasets, each containing 2,000 samples. These datasets are formatted as comma-separated value (CSV) files, featuring two essential columns: a text column and a binary label column. The binary labels, denoted as 0 and 1, signify whether the text belongs to a specific ESG category or not. The datasets include:

- environmental_2k.csv (Environmental dataset)
- social_2k.csv (Social dataset)
- governance_2k.csv (Governance dataset)

Moving on to the source code folder, it houses various notebooks:

- preprocess_esg_data.ipynb: This Jupyter notebook, scripted in Python, is dedicated to transforming the label columns within the datasets. For instance, it converts the labels in environmental_2k.csv from 0/1 to 'Not Environmental'/'Environmental'. Similar processing is applied to other datasets, with the results saved as CSV files.
- llama2: This directory encompasses three Python Jupyter notebooks aimed at fine-tuning and evaluating Llama 2 for ESG text classification. The notebooks include:
    - Fine-tune Llama 2 for Text Classification-Environmental.ipynb
    - Fine-tune Llama 2 for Text Classification-Social.ipynb
    - Fine-tune Llama 2 for Text Classification-Governance.ipynb
- Classical ML: This section holds three Python Jupyter notebooks designed for constructing SVM and XGBoost classifiers, subsequently evaluating them for ESG text classification. The notebooks are titled:
    - SVM, XGBoost Text Classification-Environmental.ipynb
    - SVM, XGBoost Text Classification-Social.ipynb
    - SVM, XGBoost Text Classification-Governance.ipynb
- Finbert-esg: Lastly, this directory contains three Python Jupyter notebooks dedicated to evaluating Finbert-esg classifiers for ESG text classification:
    - Evaluate FinBERT-ESG for Text Classification-Environmental.ipynb
    - Evaluate FinBERT-ESG for Text Classification-Social.ipynb
    - Evaluate FinBERT-ESG for Text Classification-Governance.ipynb

Together, these components form a comprehensive toolkit for conducting ESG text classification and analysis.